\documentclass[10pt,twocolumn,letterpaper]{article}

\usepackage{iccv}
\usepackage{times}
\usepackage{epsfig}
\usepackage{graphicx}
\usepackage{amsmath}
\usepackage{amssymb}
\usepackage{floatrow}

\newcommand{\blue}[1]{{\color{blue} #1}}

\usepackage[pagebackref=true,breaklinks=true,colorlinks,bookmarks=false]{hyperref}

\def\iccvPaperID{7092} 
\def\confYear{ICCV 2021}
\iccvfinalcopy 

\begin{document}


\title{Test-Time Adaptation for Super-Resolution: You Only Need to \\ Overfit on a Few More Images}

\author{Mohammad Saeed Rad \quad \quad

Thomas Yu \quad \quad

Behzad Bozorgtabar \quad \quad

Jean-Philippe Thiran
\and
Signal Processing Lab (LTS5), EPFL, Lausanne, Switzerland\\
{\tt\small \{saeed.rad, firstname.lastname\}@epfl.ch}
}

\maketitle


\begin{abstract}
Existing reference (RF)-based super-resolution (SR) models try to improve perceptual quality in SR under the assumption of the availability of high-resolution RF images paired with low-resolution (LR) inputs at testing. As the RF images should be similar in terms of content, colors, contrast, etc. to the test image, this hinders the applicability in a real scenario. Other approaches to increase the perceptual quality of images, including perceptual loss and adversarial losses, tend to dramatically decrease fidelity to the ground-truth through significant decreases in PSNR/SSIM. 
Addressing both issues, we propose a simple yet universal approach to improve the perceptual quality of the HR prediction from a pre-trained SR network on a given LR input by further fine-tuning the SR network on a subset of images from the training dataset with similar patterns of activation as the initial HR prediction, with respect to the filters of a feature extractor. In particular, we show the effects of fine-tuning on these images in terms of the perceptual quality and PSNR/SSIM values. Contrary to perceptually driven approaches, we demonstrate that the fine-tuned network produces a HR prediction with both greater perceptual quality and minimal changes to the PSNR/SSIM with respect to the initial HR prediction. Further, we present novel numerical experiments concerning the filters of SR networks, where we show through filter correlation, that the filters of the fine-tuned network from our method are closer to ``ideal'' filters, than those of the baseline network or a network fine-tuned on random images.
  
\end{abstract}

 \begin{figure}
   \centering
   \includegraphics[width=1.00\linewidth]{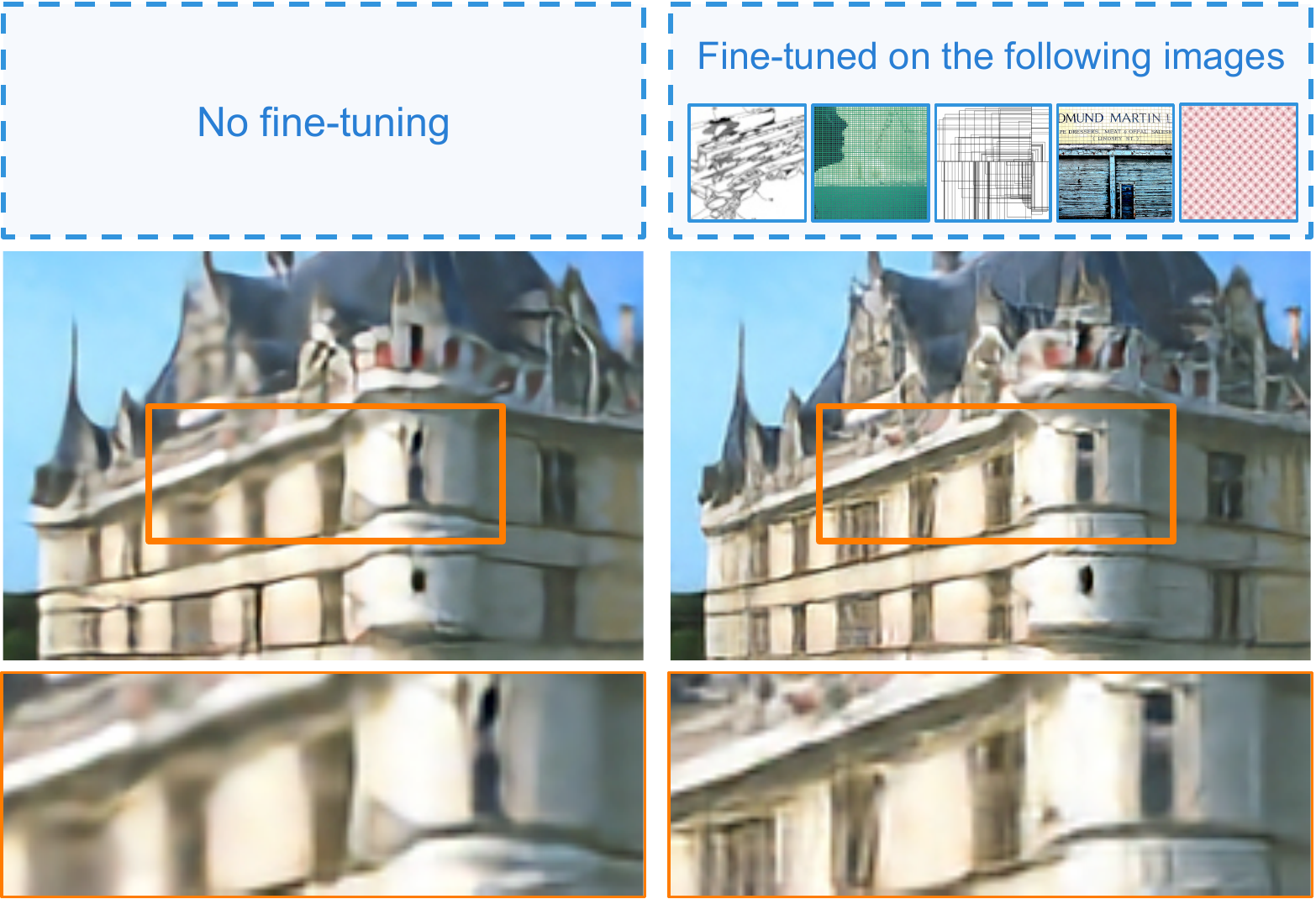}
   \caption{We demonstrate how we can improve the perceptual quality of Super-Resolution images produced by a generic SR network and a given LR image by fine-tuning the network on specific images which activate the same filters of a pre-trained feature extractor as those activated by the initial SR prediction. Left: Initial SR predictions from the baseline network, right: Predictions from the network after fine-tuning for a few iterations on selected images by our method. Zoom in for the best view.}
   \label{fig:cover_pic}
 \end{figure}

\section{Introduction}

Super-resolution (SR) is the ill-posed problem of transforming low-resolution (LR) images ($I_{LR}$) to their high-resolution (HR) counterparts ($I_{HR}$) \cite{srsurvey,wang2019deep,anwar2019deep,dong2014learning,Rad_2021_WACV,tai2017memnet}. A common way to model the interaction between LR and HR images can be formulated as $I_{LR}=(I_{HR}*\mathbf{k})\downarrow_s+ \; N$, where $*$ denotes convolution, $\mathbf{k}$ is the blur kernel, $\downarrow_s$ denotes downsampling by a factor $s$, and $N$ is noise. In this paper, we focus on a common setting for SR, where the down-sampling kernel is known and is a bicubic downscaling kernel \cite{srsurvey}.

\begin{figure*}[t]
\begin{center}
\makebox[0pt]{
\includegraphics[width=0.95\linewidth]{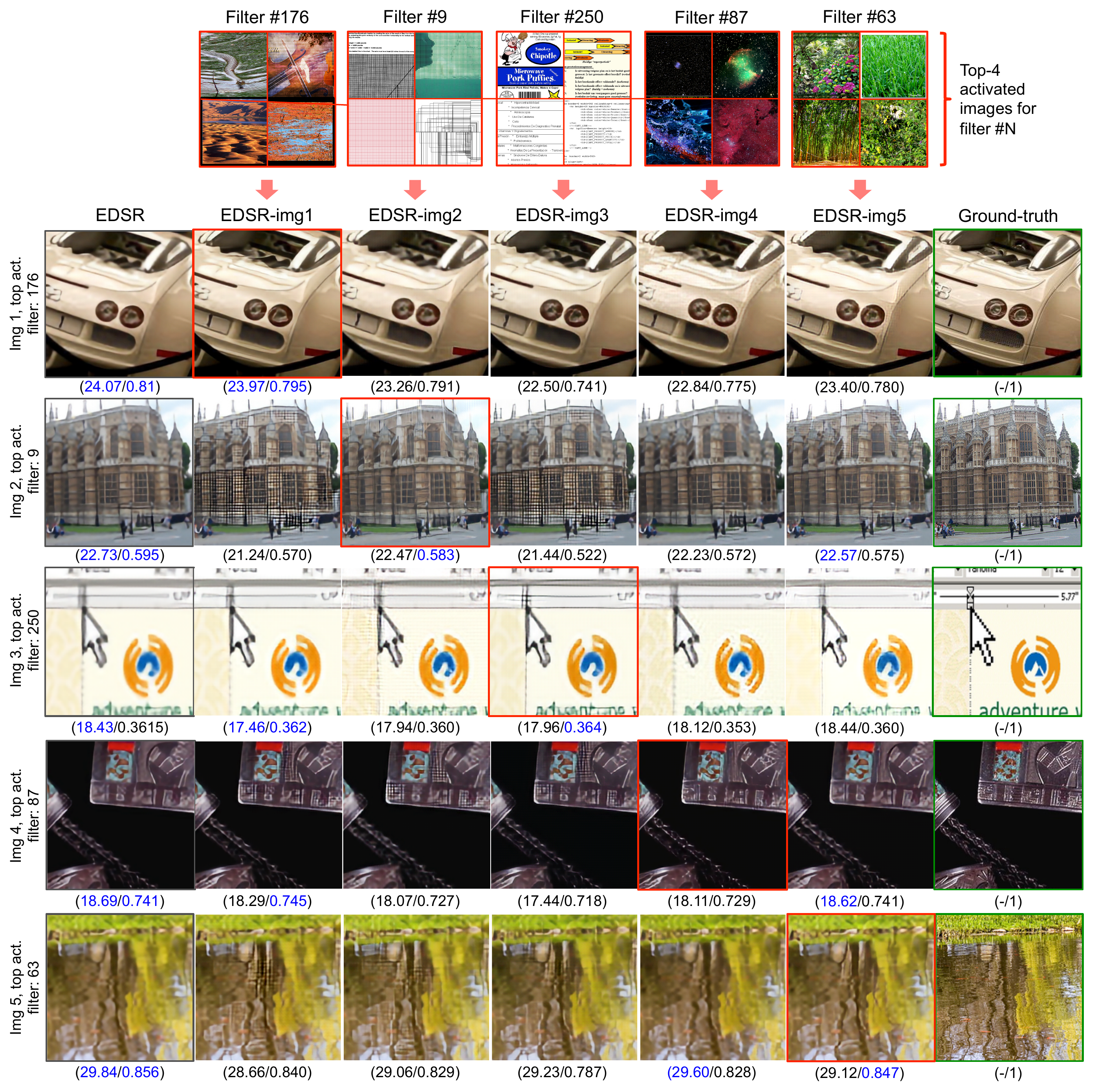}}
\end{center}
   \caption{We demonstrate the effect of fine-tuning on images, which maximally activate specific filters in a pre-trained classification network with respect to perceptual quality and PSNR/SSIM values. The first column shows the initial HR predictions from the baseline network while subsequent columns show predictions from the network after fine-tuning on the images bordered by red at the top. Note that in each row, \textbf{the network fine-tuned on the image set which shares the filter of maximal activation with the initial HR prediction gives the best perceptual quality without affecting the PSNR or SSIM significantly}. Fine-tuning on image sets which maximally activate different filters results in oversmoothing or image artifacts as compared to the ground truth. Two best values are in \blue{blue}. \textbf{Please zoom in on the screen.}}
\label{fig:qualitative_result}
\end{figure*}

In this setting, deep learning algorithms \cite{wang2018esrgan,dai2019second,zhang2018image,Rad_2019_ICCV, RAD2020304} have made remarkable progress in image super-resolution that aim to obtain a $I_{HR}$ output from one of its $I_{LR}$ versions by leveraging the power of deep convolutional neural networks. Going even further, in the field of reference (RF)-based SR, an external high-resolution reference image is provided, where the reference image and  $I_{HR}$ share similar textures and qualities \cite{zhang2019image,zheng2018crossnet,jiang2019ensemble,yang2018reference}. In this way, the networks are trained to leverage additional information from the reference HR image. \textbf{This has the drawback of assuming the existence of and finding HR images similar to a given LR image -in terms of content, colors, contrast} as well as the increased size of the networks trained to incorporate the additional HR input. 

In the SR literature, pixel-based metrics, which compare predicted HR images to the ground truth HR image such as the peak signal to noise ratio (PSNR) or structural similarity index (SSIM) are commonly used to judge the performance of SR methods \cite{srsurvey}. However, it is known that optimizing neural networks for PSNR, SSIM, or other pixel-based metrics generally result in over-smoothed, perceptually unappealing HR images \cite{blau2018perception, johnson2016perceptual, ledig2017photo}. In fact, \cite{blau2018perception} shows that there is a mathematical tradeoff between performance on these pixel-based metrics and perceptual quality. However, \textbf{we note that in theory, a perfect reconstruction would have the highest performance on both pixel-based metrics and perceptual quality}. Strategies to increase perceptual quality include training networks with a perceptual loss \cite{johnson2016perceptual}, which computes the distance between predicted and ground truth images in the feature space using a pre-trained classification network. Generative adversarial networks (GANs) \cite{goodfellow2014generative} are also used to improve perceptual quality \cite{isola2017image,ledig2017photo,wang2018esrgan,bozorgtabar2019learn,bozorgtabar2019using,sajjadi2017enhancenet,wang2018recovering}. \textbf{However, these approaches significantly decrease PSNR, SSIM and other pixel-based metrics with respect to trained networks using only the pixel-wise losses} \cite{johnson2016perceptual,ledig2017photo,blau2018perception}.


In this paper, inspired by RF-based SR and previous analysis of learned filters of classification networks, we propose a novel method to increase the perceptual quality of the output of a generic PSNR-based SR network on a given LR image without significantly affecting the PSNR or SSIM. This is done through test-time adaptation of the generic SR network, to tailor it to a given LR image used for testing. Concretely, given an input LR image and the SR network pre-trained using only pixel-wise losses, e.g., $L_1$, we first obtain the initial HR prediction from the network. We then fine-tune the network on a few pairs of LR/HR images from the training dataset, where the images are chosen by the similarity of their activations of filters from a pre-trained classification network with respect to the corresponding activations of the initial HR prediction. We show that the perceptual quality of the HR image from the fine-tuned network increases without significantly decreasing the PSNR or SSIM values. Further, we demonstrate that this does not contradict past studies on the trade-off between PSNR and perceptual quality \cite{paper_enhanced, paper_twitter_0}, as this results from fine-tuning on images that activate the same filters as the initial LR input. The fine-tuned SR network performs worse on images dissimilar to the LR input; hence, overall performance is in conformity with the trade-off. As shown in Fig. \ref{fig:qualitative_result}, our method can improve perceptual quality with minimal impact on PSNR/SSIM with fine-tuning on images with similar activations as the LR input.

Our contributions are as follows:

\begin{itemize}
    \item We propose a novel, test-time adaptation method to improve SR, which guides PSNR-based SR networks toward perceptually more compelling images by fine-tuning on selected images at the test-time, \textbf{without significant impact on the PSNR or SSIM}.
    \item To our knowledge, we are the first to investigate how overfitting/fine-tuning on selected images, which differ by what filters in a pre-trained classification network they maximally activate, can change SR reconstructions for better or worse. 
    \item We also show, to our knowledge, novel numerical experiments in the field of SR, where we quantitatively relate the filters of the pre-trained SR network, the fine-tuned network, and an ``ideal'' SR network (ideal with respect to the given LR input) to show that our method moves the filters of the pre-trained SR network closer to the ``ideal'' filters.  
\end{itemize}

\begin{figure*}[t]
\vspace{-5mm}
\begin{center}
\makebox[0pt]{
\includegraphics[width=0.7\linewidth]{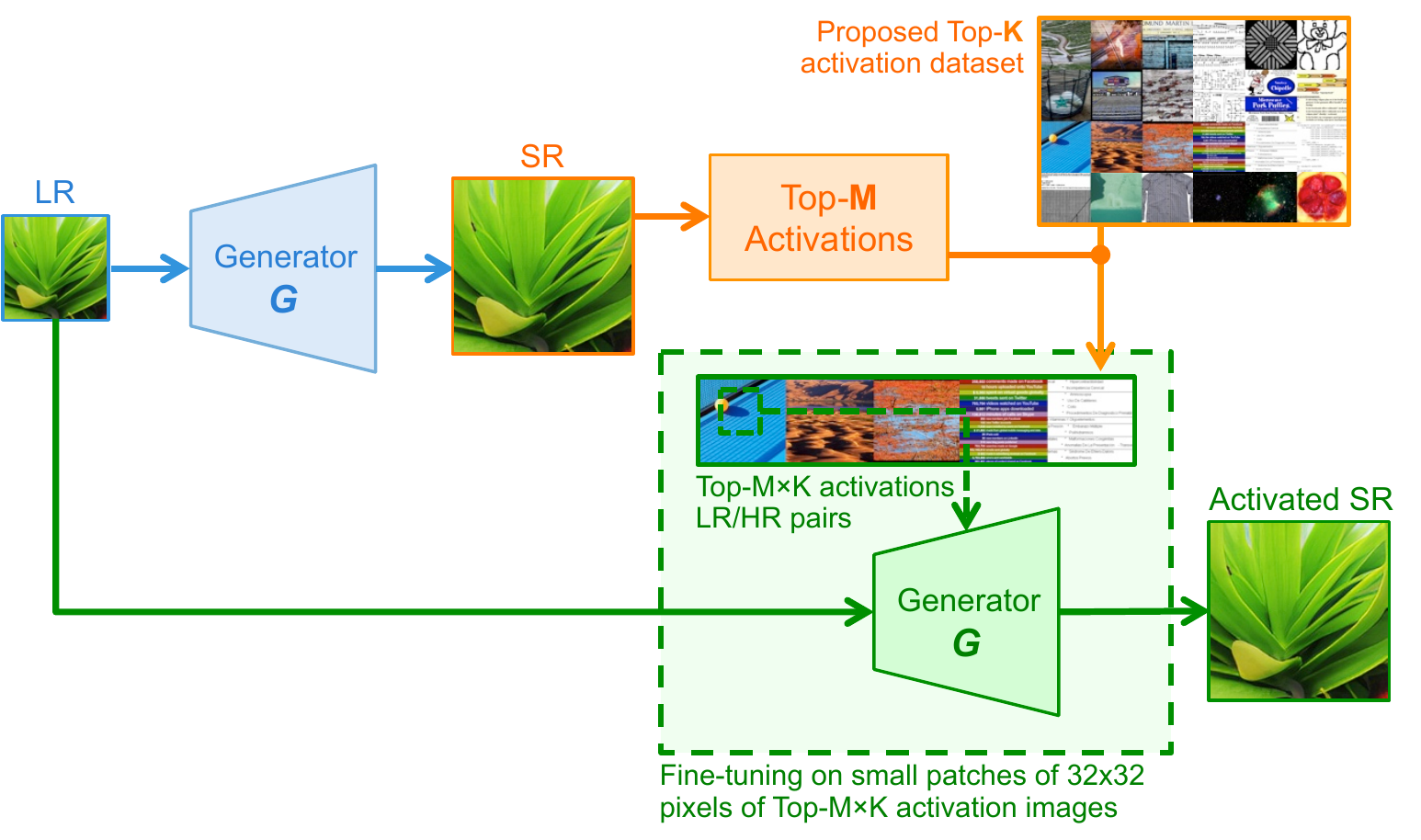}}
\end{center}
   \caption{The overview of the proposed method: First, the LR input is passed to the SR network to generate an initial SR prediction. We then find the top $M$ filters of the third layer of the VGG~\cite{paper_vgg} network which are activated by the initial SR prediction. Then, we fine-tune the SR network on a set of $M*K$ images chosen from the training data, which maximally activate the same $M$ filters. Finally, we pass the LR input to the fine-tuned SR network for the final SR prediction. In this example, K = 1 and M = 5.}
\label{fig:overview}
\end{figure*}

\section{Overview of the approach}
\label{sec:overview}
The overview of the proposed method is shown in Fig. \ref{fig:overview}; the task is to predict an HR image from a given LR input by benefiting from a few more essential images with respect to the pre-trained model. The pipeline can be split into three main steps: First, we construct a reference dataset, namely the Activation dataset, containing essential images for further fine-tuning. Second, we use a novel technique to choose relevant images from the Activation dataset. Finally, we fine-tune the pre-trained SR network on these images and produce the final reconstruction. In what follows, let $G$, $\mathcal{D}$ denote the baseline SR network and the dataset of paired LR and HR images used to train the SR network, respectively. We present each step in detail as follows:


\noindent
\textbf{Construction of Activation dataset}
We first construct a reference dataset from the HR images of $\mathcal{D}$ by extracting their corresponding activations from the third layer of the VGG classification network \cite{paper_vgg}. For each channel in the third layer ($conv3$), we order (descending) the images by the channel's corresponding activation and take the top $K$ images. As there are 256 channels in the third layer, we form a reference dataset of $256 \times K$ HR images. We choose the third layer as the features from this layer have been shown to be more discriminative \cite{caron2018deep,zhang2017split}. As an example, in Fig. \ref{fig:topacts}, we show for different filters in different layers of VGG19~\cite{paper_vgg}, the top nine images by filter activation from a subset of 50 thousand images from ImageNet~\cite{deng2009imagenet}. We further investigate the effectiveness of using other layers ($conv2, 4$, and $5$, in our supplementary material).


\noindent
\textbf{Test-Time Adaptation of the SR network}
We obtain an initial HR prediction from passing LR to $G$, which we call $SR$. We pass $SR$ through the third layer of the VGG classification network \cite{paper_vgg} and note the top $M$ filters with the highest activations. From this list of filters, we can use our reference dataset to define a set of $M \times K$ images where for each of the $M$ filters, we take the top $K$ images in our dataset in terms of activation of the filter. We then fine-tune $G$ on this set of images for a set number of epochs determined by performance on the validation set. 

\noindent
\textbf{Prediction}
After fine-tuning $G$, we again pass the LR image to $G$ to obtain our final HR prediction, which we call activated SR. \textbf{The activated SR image is perceptually more convincing than the initial SR, without significant decreases in its PSNR and SSIM values.}

\section{Image Activations in SR}
\begin{figure*}[h]
\vspace{-5mm}
\begin{center}
\makebox[0pt]{
\includegraphics[width=0.8\linewidth]{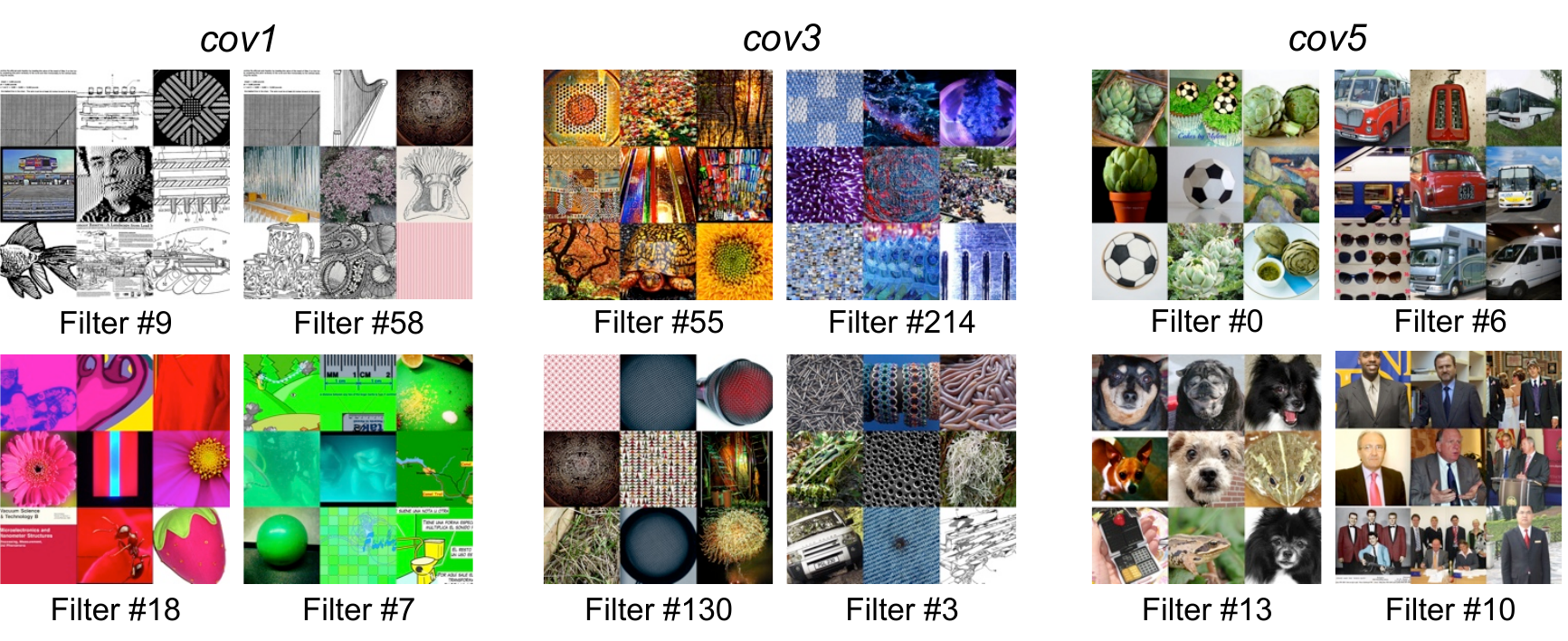}}
\end{center}
\vspace{-3mm}
   \caption{Top 9 activated images from a subset of 50 thousand images from ImageNet~\cite{deng2009imagenet} for different filters in the conv1, conv3 and conv5 layers of VGG19~\cite{paper_vgg}, respectively.
}
\label{fig:topacts}
\end{figure*}

In the machine learning/computer vision literature, analysis of the activations of neural networks with respect to different inputs is often used for the purposes of understanding/interpretability \cite{bau2017network} and extraction of relevant features for downstream processing, for instance, in unsupervised learning \cite{caron2018deep, simon2015neural}. In terms of SR, only perceptual loss uses this analysis by matching the activations, with respect to a layer of a pre-trained classification network, of the HR prediction and the ground-truth, showing the efficacy and importance of these features. We go further by explicitly analyzing the activations of the third layer of VGG19~\cite{paper_vgg} with respect to a large dataset of 50 thousand images. Then for each filter, we can assign a group of images with the highest activations. As perceptual loss shows that constructing images based on activations can improve perceptual quality, it stands to reason that fine-tuning a network on images that also triggers specific filters can enhance SR reconstructions on images that have similar activations with respect to those filters. Hence, in contrast to perceptual loss, we are able to exploit the analysis of activations by enhancing the perceptual quality of SR on a given LR input by using a set of images \textbf{which are visually different from the LR input}, but similar in terms of activation. To the best of our knowledge, we are the first to create and benefit from such a dataset for SR. Fig. \ref{fig:topacts} shows a few example images from the Activated dataset; the detailed procedure of generating this dataset is presented in section \ref{sec:overview}. \textit{This dataset is available in supplementary material and we will release this dataset upon acceptance of the paper}.


\section{Overfitting: the good, the bad, and the ugly}
\begin{figure*}[t]
\vspace{-3mm}
\begin{center}
\makebox[0pt]{
\includegraphics[width=1.0\linewidth]{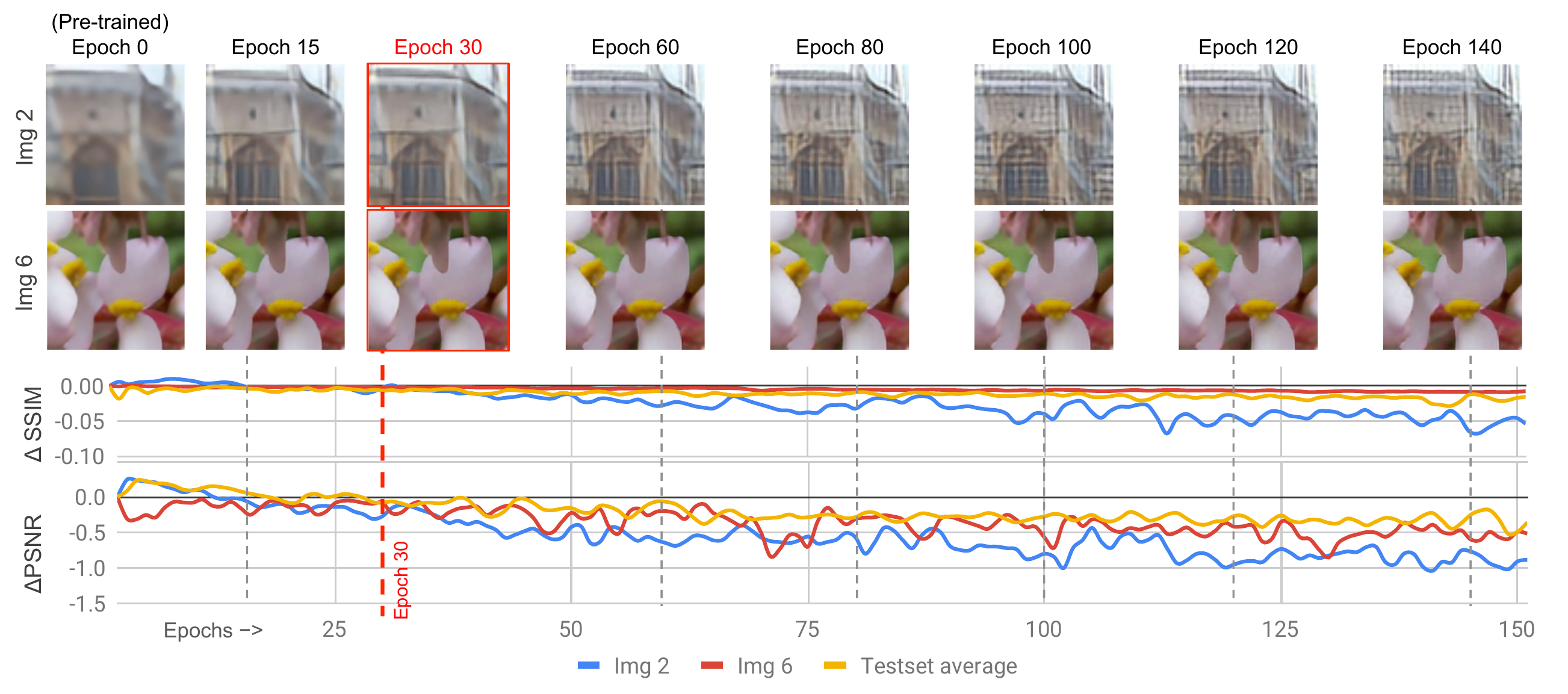}}
\end{center}
   \caption{The effects of fine-tuning as a function of the number of epochs. We show the average change of PSNR and SSIM values over the test set, as well as explicit examples of visual, PSNR, and SSIM evolution on two images. We see in image 2 that perceptual quality can dramatically increase with fine-tuning, while image 6 is not affected significantly. \textbf{Please zoom in on the screen.}}
\label{fig:qualitative_result_1}
\end{figure*}

Throughout this paper, we have used the word ``fine-tuning'' for continuing the training of a pre-trained SR network on a small set of images. Implicitly, this assumes that such training has a beneficial effect for the purpose of the network, which is to perform SR on a given LR image (\textbf{``The good''}). However, as seen in Fig.~\ref{fig:qualitative_result}, such fine-tuning could also be labeled as overfitting, since our method only improves reconstructions on images with similar patterns of filter activation as the given LR image; other inputs can result in image artifacts and over smoothing (\textbf{``The bad''}). That is, the fine-tuned network no longer generalizes to all image classes. This can be understood in terms of the tradeoff between perceptual quality, and PSNR established in \cite{blau2018perception}. We conjecture that we are able to gain perceptual quality with minimal changes to PSNR/SSIM precisely because this gain occurs only on images similar in filter activation to those used in the fine-tuning. As both PSNR and perceptual quality can decrease in other images, the overall performance does not contravene the tradeoff. Thus, for a given LR image, overfitting is actually good for improving SR reconstructions. However, we note that the outcome of fine-tuning is dependent on the number of epochs of additional training (\textbf{``The ugly''}). Further, while generalization of the network performance is clearly compromised, it is possible for the fine-tuning to have no effect, good or bad, on different classes of images. In Fig. \ref{fig:qualitative_result_1}, we show the effects of fine-tuning on visual quality, PSNR, and SSIM values as a function of the number of epochs as well as how it can dramatically increase the perceptual quality of some images while not affecting others. 

\begin{figure*}[t]
\begin{center}
\makebox[0pt]{
\includegraphics[width=0.8\linewidth]{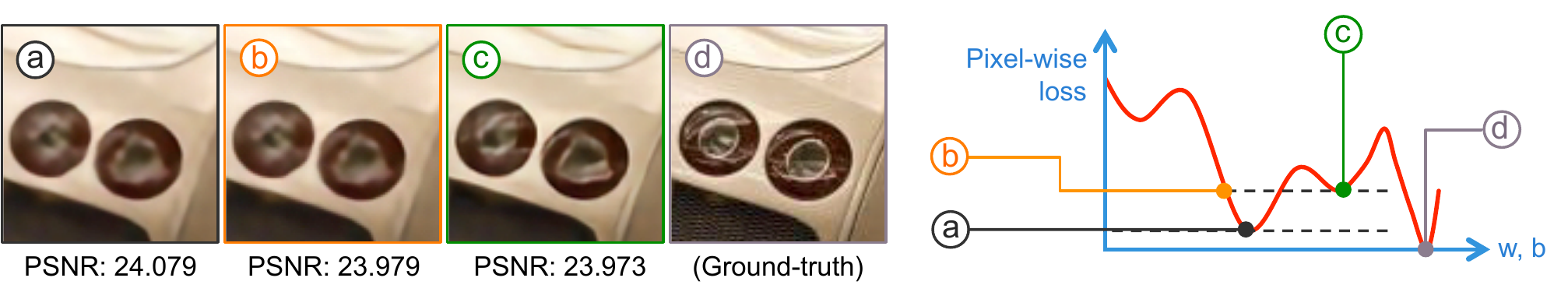}}
\end{center}
   \caption{Image (a) is obtained by a pre-trained baseline with pixel-wise optimization on a large dataset. Images (b,c) are obtained during the fine-tuning by our proposed method, reaching almost the same PSNR. Image (d) is the ground truth. We see from comparing images (b) and (c) that our method is guiding the SR network to a different local minimum with a better perceptual quality, as the same loss is achieved but with dramatically different quality.}
\label{fig:qualitative_result_2}
\end{figure*}

It remains to address how overfitting using only a pixel-wise loss can improve perceptual quality. We emphasize that the fine-tuning is done with only $L_1$ loss; in contrast to perceptual loss or adversarial losses used to improve perceptual quality, only pixel-wise metrics are used in our approach. In Fig. \ref{fig:qualitative_result_2}, we show a diagram of our hypothesis that 
\textbf{overfitting guides the SR network to a local minimum, where the pixel-wise error is only slightly different, while the perceptual quality is dramatically improved.} As evidence, note that almost the same PSNR is achieved on image b (during the pretraining of the network, before fine-tuning by our approach) and image c (after fine-tuning), but image c is much sharper and realistic. 

\section{Experiments and results}
\subsection{Experimental settings}
\subsubsection{Generator architecture}
While our method and experiments can generalize to arbitrary SR networks, we use an  EDSR~\cite{paper_edsr} as our baseline generator, which we denote as $G$. EDSR performs better than other conventional residual SR networks by eliminating some unnecessary modules e.g., batch normalization. This makes it a good candidate to investigate the effectiveness of our proposed approach as many other SR networks incorporate components designed for specific contributions/improvements that may not strictly be necessary. The architecture consists of 32 residual blocks and 256 filters per convolutional layer (more details in supplementary material). We train this network in a single step for 50 epochs, using the $L_1$ loss function. For the training data, we use a subset of 50 thousand images taken from Imagenet~\cite{deng2009imagenet}. The Adam optimizer was used for the optimization. The learning rate was set to $1e{-3}$ and then decayed by a factor of ten every 20 epochs.

\subsubsection{Fine-tuning/overfitting}
\noindent
\textbf{Parameters: }
In order to force the fine-tuning to make changes to the filters of the network' feature extractor rather than changing the last layers of the network, we freeze the convolutional layers related to up-sampling, more specifically, the filters coming after the pixel-shuffle layers. The images for fine-tuning are the random crops of $32 \times 32$ pixels from our constructed dataset. We choose a relatively low learning rate of $1e-4$ for gradual change.

\noindent
\textbf{K and M: }
We conduct sensitivity analysis to choose the best values for the number of images per filter $K$ and the number of filter $M$ used for our test image. We tune these parameters based on the perceptual quality of the generated images. The results of this work are produced by setting the values of $K$ and $M$ to two and five, respectively (10 images in total). a more detailed study can be found in the supplementary material.  

\noindent
\textbf{Stoppage condition: }
The criteria to stop the fine-tuning was basically defined based on qualitative comparison of reconstructed images at different epochs where we could see at epoch 30, the vast majority of the images from our validation set were perceptually more convincing as compared to other epochs. However, considering Fig.~\ref{fig:qualitative_result_2}, we can see this choice can also be justified as this epoch also coincides with the beginning of a significant drop in SSIM and PSNR values over all images on the test set.

\subsubsection{Test-set}
For our test-set, we randomly chose 100 images from the ImageNet dataset (non-overlapping between activation and training datasets), as both our baseline network and the Activation dataset are trained on/using a subset of 50,000 ImageNet images. As it is shown \cite{Han2019unsupervised,wei2020unsupervised} that SR network quality drops when doing cross-dataset tests, therefore, we focus on showing a proof of concept of improving a generic SR network on a generic dataset and do not add an additional variable of different datasets to the mix.

\begin{figure*}[t]
\vspace{-3mm}
\begin{center}
\makebox[0pt]{
\includegraphics[width=0.95\linewidth]{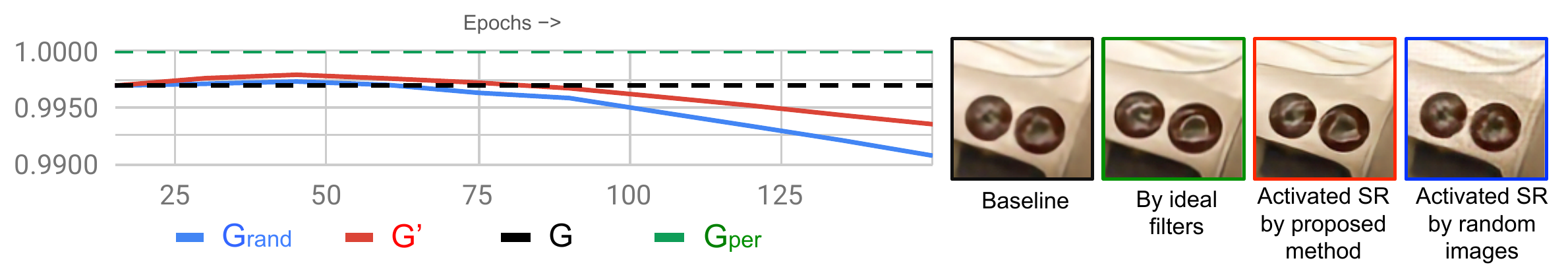}}
\end{center}
   \caption{The average correlations over the test-set images of the filters of the final layer of feature extractor of $G'$ and $G_{rand}$ to the filters of $G_{per}$ as a function of the number of epochs of fine-tuning in red/blue respectively, with the correlation of the baseline as a dotted black line. We see that the correlation of $G'$ to $G_{per}$ is higher than $G_{rand}$; This is consistent with our hypothesis that the proposed method of fine-tuning transforms the filters of the baseline to be closer to the ``ideal'' filters for a particular image.} 
\label{fig:correlation}
\vspace{-4mm}
\end{figure*}

\subsection{Filter selection analysis}
In the following, we provide, to our knowledge, novel experiments and investigations into SR networks, where we examine, at the level of the network' filters, how the SR network changes in response to our selective overfitting. For our experiments, we draw on \cite{wang2019deep_2}, 
where authors found that two networks trained from scratch for the same task can have different filter orders and different filter patterns; however, fine-tuning a network to perform a different, but related task preserved the filter orders and patterns of the original network. They further show that the changes in filters by doing fine-tuning are gradual, by proposing to quantitatively assess the similarities between the filters of two different instances of the same network through correlation; concretely, given filter $F_i$, $F_j$, 
\begin{align}
    \rho_{ij} = \frac{(F_i-\overline{F_i})(F_j-\overline{F_j})}{\sqrt{\| F_i-\overline{F_i}\|_2}\sqrt{\| F_j-\overline{F_j}\|_2}}
\end{align}
where $\rho_{ij}$ is the correlation index. We use this correlation index to quantitatively study the changes in the filters of the SR network after fine-tuning. Given an LR image with HR ground truth, let $G_{per}$ denote the EDSR baseline which is fine-tuned on solely this LR image to produce a perfect reconstruction. We can, in some sense, assume that $G_{per}$ possesses the ideal or optimal set of filters for super-resolving this LR image, as we overfit it on this image; further, we verified that, consistent with \cite{wang2019deep_2}, the overall structure/filter orders are preserved from the baseline network, indicating that $G_{per}$ is not simply memorizing the image within its parameters. 

Let $G'$ denote the fine-tuned network produced from our method on this LR image. Let $G_{rand}$ denote the EDSR network fine-tuned on a set of random images. In Fig \ref{fig:correlation}, we show the average correlations of the filters of the final layer of $G'$ and $G_{rand}$ to the filters of $G_{per}$ as a function of the number of epochs of fine-tuning. The average was computed by constructing $G',G_{rand},G_{per}$ for each image in the test set, then taking the average correlation over the images. We also show the correlation of the filters of the baseline $G$ with $G_{per}$.  We see that the correlation of $G'$ to $G_{per}$ is generally higher than those of $G_{rand}$ and $G$, including at 30 epochs, which is the number that we use for our method. This provides evidence that our method of fine-tuning in some sense brings the baseline closer to the "ideal" set of filters for a given LR image. 

 \begin{figure*}
 \vspace{-6mm}
   \centering
   \includegraphics[width=0.85\linewidth]{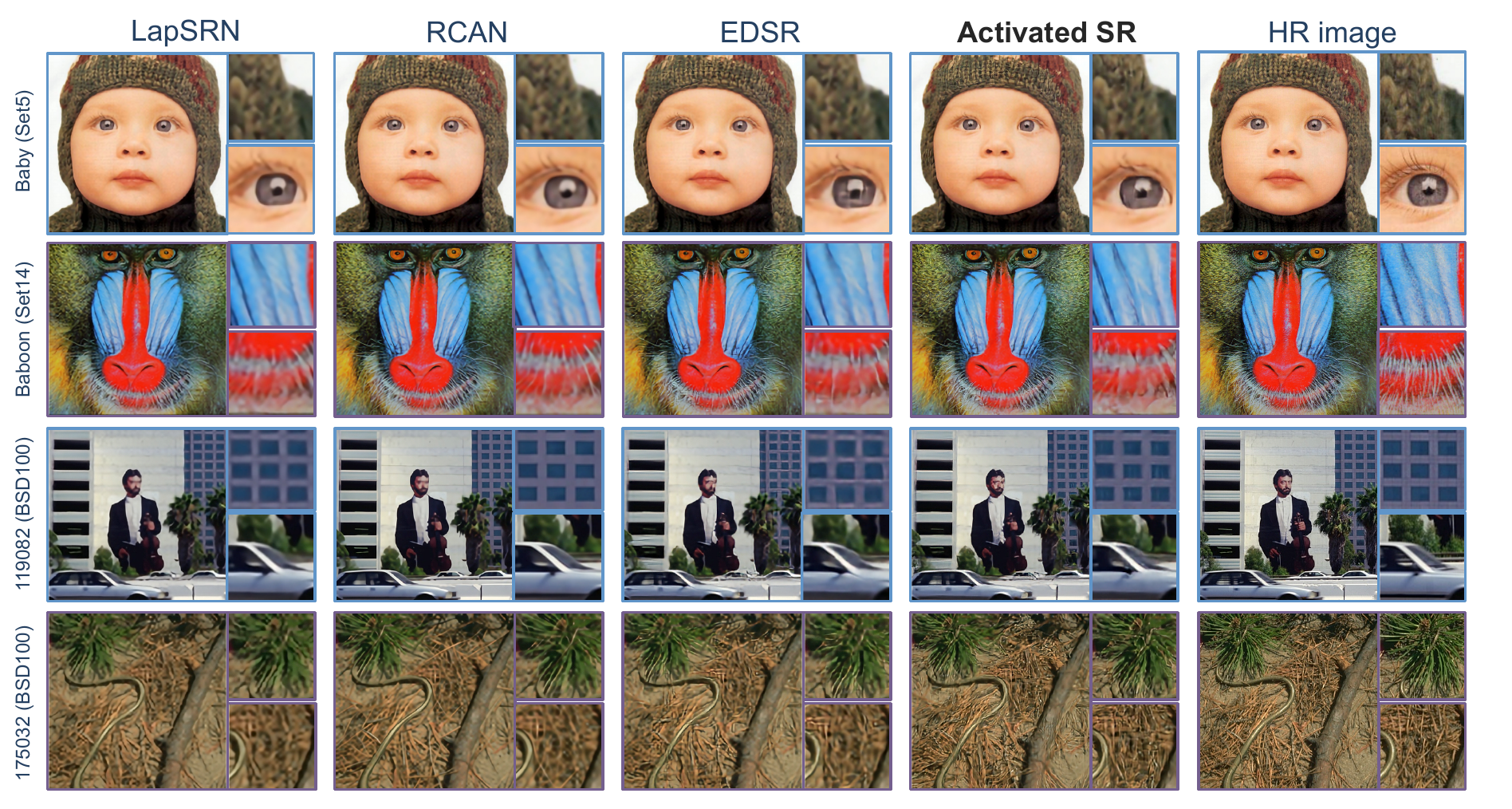}
   \caption{Qualitative comparison to \textbf{PSNR-based approaches}. From left to right: Bicubic, LapSRN~\cite{paper_lapsrn}, RCAN~\cite{paper_rcan}, EDSR~\cite{paper_edsr}, Activated-SR (ours), and HR image, tested on images from Set 5~\cite{paper_set5}, Set14~\cite{paper_set14} and BSD100~\cite{bsd100_paper} testsets. \textbf{We emphasize that our method is EDSR using our test-time adapation method.} We show results from other networks for comparison. Zoom in for the best view.  \vspace{-1mm}}
   \label{fig:comparision_pnsr_based}
    \vspace{-1mm}
 \end{figure*}
 
\subsection{Comparison to PSNR-based approaches}
From the qualitative results in Fig.~\ref{fig:qualitative_result}, we can observe that when we fine-tune the pre-trained EDSR network using the images chosen through our method, namely activated-SR approach, \textbf{the perceptual quality increases with minimal impact on the PSNR/SSIM}. This minimal impact on the PSNR/SSIM has been also shown in Fig.~\ref{fig:qualitative_result_1}, where we can see that over a test set of 100 images, the mean changes in PSNR/SSIM are minimal.

In Fig.~\ref{fig:comparision_pnsr_based}, we additionally compare our method to LapSRN~\cite{paper_lapsrn}, RCAN~\cite{paper_rcan} and EDSR~\cite{paper_edsr} methods and by using test images from Set5~\cite{paper_set5}, Set14~\cite{paper_set14} and BSD100~\cite{bsd100_paper} standard datasets. \textbf{For a fair comparison, in this section, we only considered PSNR-based approaches as our methods still relies only on minimizing the pixel-wise distance of the SR and ground-truth images and does not benefit from any perceptual losses}. This figures shows that activated-SR images produced by out method have superior perceptual quality, while Table~\ref{tab:psnr_ssim} confirms that this increases had a minimal impact on the PSNR/SSIM over the whole test set.

\begin{table}[h]

\begin{center}
\begin{tabular}{ c | c | c c c c }
\small \\[-1em]
 \small Dataset & \small Metric & \small LapSRN & \small RCAN & \small EDSR & \small \textbf{Ours}\\
\hline \\[-1em]
\small Set5 & \small SSIM & 0.887 & \textbf{0.918}  & 0.893 & 0.891 \\ 
\small \space & \small PSNR & 31.56 & \textbf{32.61}  & 32.41 & 32.40 \\
\hline \\[-1em]
\small Set14 & \small SSIM & 0.772 & 0.773 & 0.774 & \textbf{0.776} \\ 
\small \space & \small PSNR & 28.20 & \textbf{28.86}  & 28.81 & 28.70 \\
\hline \\[-1em]
\small BSD100 & \small SSIM & 0.742 & 0.815 & 0.802 & \textbf{0.819} \\ 
\small \space & \small PSNR & 27.41 & \textbf{29.32}  & 29.24 & 29.15 \\
\end{tabular}
\end{center}
\vspace{-2mm}
\caption{
Comparison LapSRN~\cite{paper_lapsrn}, RCAN~\cite{paper_rcan}, EDSR~\cite{paper_edsr}, and activated-SR (ours) on various test sets. \textbf{We emphasize that our method is EDSR using our test-time adapation method.} We show the results from other methods for comparison. Considering Fig.~\ref{fig:comparision_pnsr_based} the proposed method improves the perceptual quality of EDSR with minimal impact on the PSNR/SSIM.\vspace{-4mm}}
\label{tab:psnr_ssim}
\vspace{-3mm}
\end{table}

\subsection{Comparison to perceptual-based approaches}
\begin{figure*}
\begin{center}
\makebox[0pt]{
\includegraphics[width=0.85\linewidth]{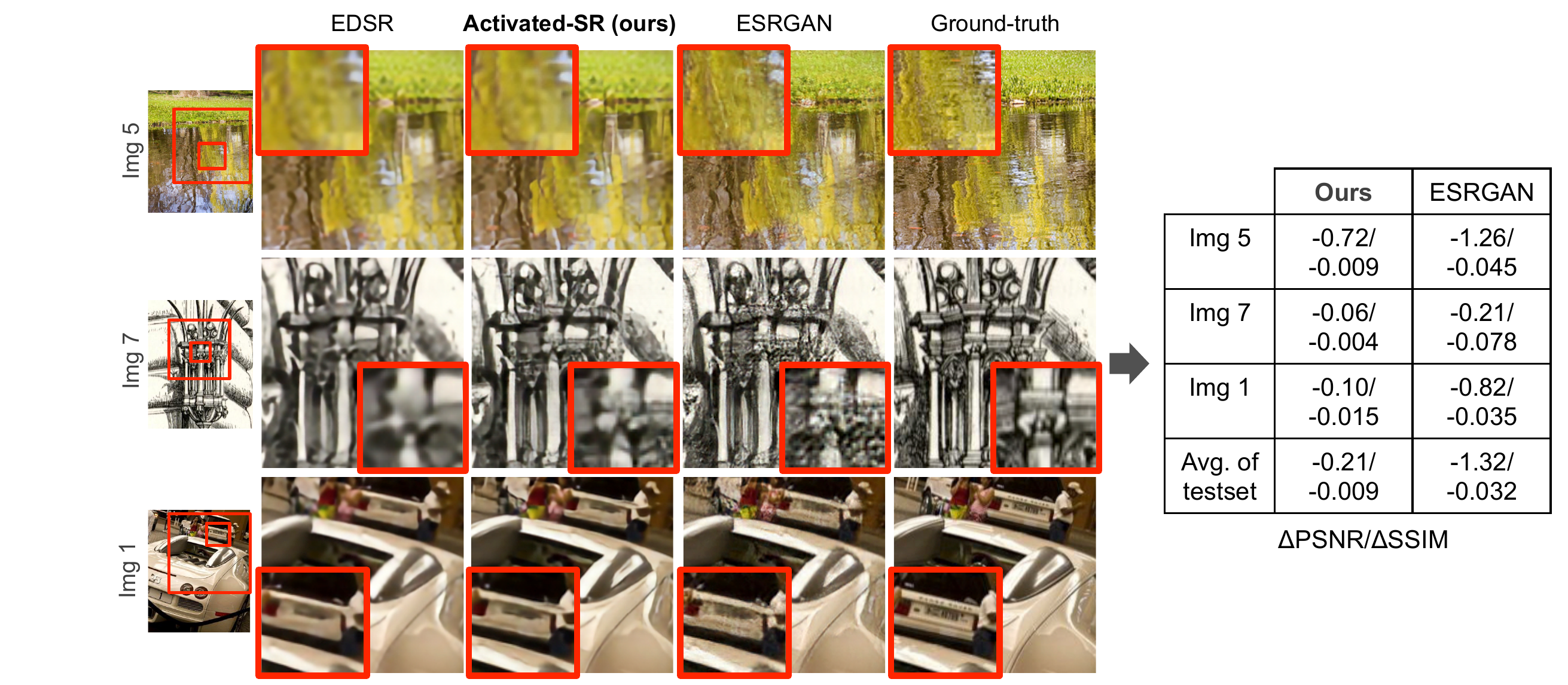}}
\end{center}
\vspace{-1mm}
   \caption{Comparing the proposed method and a perceptual-based approach \cite{paper_esrgan}. In general, the perceptual loss provides sharper edges but also more distorted textures, wheres the proposed method provides images which are sharper and contain more details than the baseline without distortion. In the table, we show that this is reflected in the decrease in the PSNR/SSIM; \bf{using perceptual loss decreases the PSNR/SSIM relative to the baseline far more than using our method.\vspace{-2mm}}}
\label{fig:comp_perc}
\vspace{-4mm}
\end{figure*}

Finally, in Fig~\ref{fig:comp_perc}, we provide a comparison between SR network trained using our proposed method and using perceptual losses (pixel-wise loss + vgg loss + adversarial loss, with the same setting and discriminator as described in ESRGAN~\cite{paper_esrgan} work). We note that the perceptual loss adds more sharpness than that of our method, but can also provide highly distorted textures. In all cases, the images from our method are sharper/more detailed than those of the EDSR baseline, without distorting the texture. This can be explained by the fact that optimizing SR networks with only perceptual loss sometimes leads to the incitement of high frequency details in image e.g., sharp edges, entailing over-sharpened images. Therefore, they do not conform with the distortion based metrics.

On average, the decrease in PSNR and SSIM using perceptual loss is 628 and 355 percent larger, respectively, than the corresponding decreases using our method. Hence, our method provides images with much greater fidelity to the ground truth, while increasing the perceptual quality without distorted textures.

\subsection{Inference time}
We note that as our method fine-tunes the baseline network for every test image, this is computationally more expensive than simply using the baseline network. However, we note that relatively small patches of $32 \times 32$ pixels, and a small number of images (10 in our case) used for fine-tuning still keeps the computation time practical for single image SR tasks; the additional fine-tuning takes $\sim$13 seconds by using a GeForce GTX 1080Ti GPU, which results in a total time of $\sim$14 seconds for a $2560 \times 1920$ pixel output.

\vspace{-2mm}
\section{Conclusion}
\vspace{-1mm}

In this paper, we propose a novel approach to improve the perceptual quality of PSNR-based SR methods. In our approach, given a pre-trained SR network and LR input, we use test-time adaptation by fine-tuning the SR network on a subset of images from the training dataset with similar activation patterns as the initial HR prediction, with respect to the filters of a feature extractor. We show that the fine-tuned network produces an HR prediction with both greater perceptual quality and minimal changes to the PSNR/SSIM, in contrast to perceptually driven approaches. Further, in contrast to reference-based SR, we use only images from our proposed activation dataset for fine-tuning, eliminating the issue with the availability of HR reference images close to the input image. Finally, through numerical experiments novel to the field of SR, we show that our fine-tuning can be interpreted as within the test-time adaptation paradigm, where we update the model parameters to be closer to the parameters of an "ideal" SR network, which is overfitted on the given LR input.

{\small
\bibliographystyle{ieee_fullname}
\bibliography{egbib}
}

\end{document}